\documentclass{article}

\usepackage{microtype}
\usepackage{graphicx}
\usepackage{subfigure}
\usepackage{booktabs} 

\usepackage{amssymb}
\usepackage{amsmath}
\usepackage{xcolor}
\usepackage{bm}
\usepackage{hyperref}

\usepackage[accepted]{icml2021}

\icmltitlerunning{UniSpeech}

\begin{document}

\twocolumn[
\icmltitle{UniSpeech: Unified Speech Representation Learning \\ with Labeled and Unlabeled Data}



\icmlsetsymbol{equal}{*}

\begin{icmlauthorlist}
\icmlauthor{Chengyi Wang}{equal,n}
\icmlauthor{Yu Wu}{ms}
\icmlauthor{Yao Qian}{ms}
\icmlauthor{Kenichi Kumatani}{ms}
\icmlauthor{Shujie Liu}{ms}
\icmlauthor{Furu Wei}{ms}
\icmlauthor{Michael Zeng}{ms}
\icmlauthor{Xuedong Huang}{ms}
\end{icmlauthorlist}
\icmlaffiliation{n}{Nankai University, Tianjin, China}
\icmlaffiliation{ms}{Microsoft}

\icmlcorrespondingauthor{Chengyi Wang}{cywang@mail.nankai.edu.cn}
\icmlcorrespondingauthor{Yu Wu}{yuwu1@microsoft.com}
\icmlcorrespondingauthor{Yao Qian}{yaoqian@microsoft.com}

\icmlkeywords{Speech recognition, Speech representation learning}

\vskip 0.3in
]



\printAffiliationsAndNotice{Work done during internship at Microsoft. \\} 

\begin{abstract}
In this paper, we propose a unified pre-training approach called UniSpeech to learn speech representations with both labeled and unlabeled data, in which supervised phonetic CTC learning and phonetically-aware contrastive self-supervised learning are conducted in a multitask learning manner. The resultant representations can capture information more correlated with phonetic structures and improve the generalization across languages and domains.
We evaluate the effectiveness of UniSpeech for cross-lingual representation learning on the public CommonVoice corpus. The results show that UniSpeech outperforms self-supervised pre-training and supervised transfer learning for speech recognition by up to 13.4\% and 26.9\% relative phone error rate respectively  (averaged over all testing languages). The transferability of UniSpeech is also verified on a domain-shift speech recognition task, demonstrating a relative word error rate reduction of 6\% against the previous approach \footnote{Code and model are available at \url{https://github.com/cywang97/unispeech}}. 
\end{abstract}

\section{Introduction}

In the past several decades, enormous progress has been made by the speech recognition (SR) community. SR systems have achieved remarkable quality and even reach human parity in many domains \cite{watanabe2018espnet, li2020comparison,wang2020transformer,xiong2016achieving,DBLP:conf/icassp/ChiuSWPNCKWRGJL18}. Unfortunately, the successful techniques require many thousands of hours of human-annotated speech recordings for training, which is not available for the vast majority of the nearly 7000 languages spoken worldwide \cite{katzner2002languages}. This poses a real challenge for building an accurate and robust SR system for low-resource languages. Even for the rich-resource languages, lack of training data is also a serious problem for specific domains, especially when the background noise and distortion conditions vary greatly from the general domain \cite{li2014overview}.

Current works tackle the low-resource speech recognition in either supervised or unsupervised manners. In the supervised case, transfer learning methods learn features on large, high-resource datasets and directly use them in similar but data-poor tasks \cite{ghoshal2013multilingual}. Though effective, it requires massive supervised corpora and neglects the large scale unlabeled data. In contrast, the unsupervised method attempts to learn powerful contextual representations from audio alone and then fine-tune the model on labeled data. For instance, wav2vec 2.0 \cite{DBLP:conf/nips/BaevskiZMA20} demonstrates a remarkable performance with 60k hours of unpaired data and 1 hour labeled data on Librispeech. \citet{DBLP:journals/corr/abs-2006-13979} further extend the model to a  multilingual setting and reduce the phoneme error rate (PER) significantly. The model jointly learns contextual speech representations and a discrete codebook of latent representations, which serves to train the model with a contrastive loss.  However, the self-supervised paradigm needs to be carefully designed and such representations may be difficult to interpret. There is no guarantee that the model learns ``good" speech representations in terms of the most valuable information for recognition. 

In the most of cases, it is less challenging to obtain labeled high-resource data and unlabeled low-resource data, while the labeled low-resource data is hard to collect. Our goal is to leverage all accessible data to learn robust representation across different languages or domains, which is capable of capturing SR-specific content, e.g. phoneme identities, while being invariant to confounding details like the background noise. With such representation, limited amounts of labeled data is sufficient to achieve acceptable performance. 

In this work, we propose a unified approach, named UniSpeech, to learn phonetically-aware contextual representations. We follow the model structure of wav2vec2.0 which consists of a feature extractor to extract latent speech representations, a Transformer context network to learn contextual representations and a quantizer to discrete latent representations. We first pre-train the model on the labeled high-resource data and unlabeled low-resource data. Then we freeze the feature extractor and fine-tune the Transformer part on a small amount of labeled low-resource data. For pre-training, we use a multitask learning manner. For labeled data, we train the model towards two objectives: the first is a sequence-level CTC loss \cite{graves2006connectionist} applied to phoneme labels for phonetic representation learning; the second is a contrastive task defined over the masked contextual representations and the discrete latent representations as in wav2vec2.0. The CTC loss aligns each contextual representation with a phoneme label. Meanwhile, the contrastive loss implicitly closes the distance between discrete representations and contextual representations, with the hope that each codeword from the codebook can also be aligned with a meaningful phoneme unit. However, this simple loss combination method leads to limited improvements. Besides, in contrastive learning, the quantizer is prone to collapse problom where only a small portion of codewords are used. And sometimes it results in locally-optimal codebooks to enable a good contrastive loss, like Voice Activity Decection coodbook or temporally invariant coodbook \cite{DBLP:journals/corr/abs-2103-08393}. Thus, we go further to explicitly guide the quantizer to learn SR-specific information. Specifically, we randomly replace a proportion of the contextual representations with quantized latent representations in the corresponding time steps and calculate the CTC loss upon the mixed representations. In our experiment, we find this method can activate more codewords and helps for learning a phonetic-aware codebook.  For those unlabeled data from low-resource setting, we only conduct contrastive learning. As the codebook is already located in the phonetic level, the model is easily adapted to the target domain. 

We evaluate the proposed method on cross-lingual SR and domain transfer tasks. On the CommonVoice dataset \cite{ardila2019common}, our UniSpeech outperforms both supervised transfer learning and unsupervised contrastive learning by a large margin in three settings: English to single low-resource language setting (one-to-one), multi-lingual high resource languages to single low-resource language setting (many-to-one), and multi-lingual high resource languages to multi-lingual low-resource language setting (many-to-many).   
In addition, we test the domain transferability from the Librispeech \cite{panayotov2015librispeech} to the Tedlium3 dataset \cite{hernandez2018ted}, where the source domain and target domain are audiobook reading and live presentation, respectively. UniSpeech achieves a relative 6$\%$ word error rate reduction against the baseline.

The main contributions of this paper can be summarized as three-folds: First, we provide a paradigm to use both labeled and unlabeled data to improve the SR performance in low-resource scenario. To the best of our knowledge, it is the first attempt to pre-train speech encoder with both supervised method and self-supervised method in a multitask learning manner; Second, we propose a new learning strategy to explicitly align the discrete latent representation to linguistic units, resulting in a meaningful speech codebook; Third, our approach significantly outperforms both self-supervised learning and supervised learning, and achieves state-of-the-art performance on CommonVoice dataset without the help of extra data.

\begin{figure*}
    \centering
    \includegraphics[width=0.9 \textwidth]{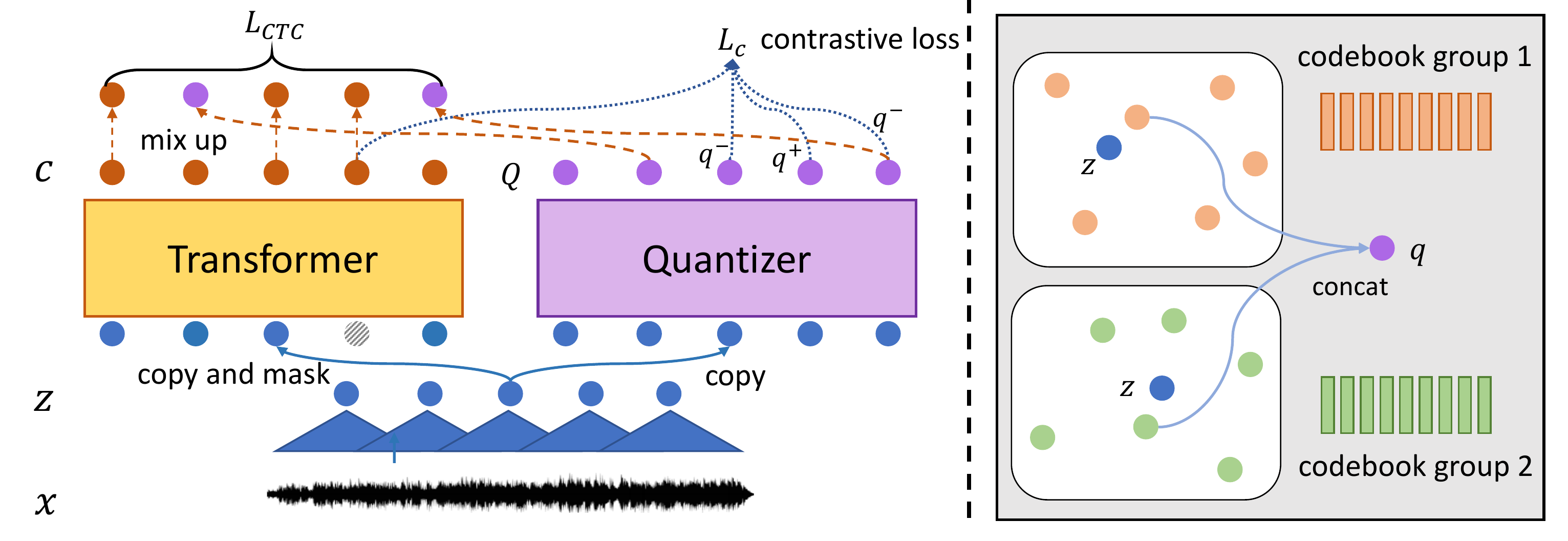}
  \caption{Left: The UniSpeech Framework. The input features to Transformer model are randomly masked following the settings in \cite{DBLP:conf/nips/BaevskiZMA20}, while the features are unmasked when fed to the quantizer layer. Right: Visualisation of the quantizer. The output of the encoder $z$ is mapped to the nearest point from the codebook.  }
  \label{model}
  \vspace{-2mm}
\end{figure*}

\section{Related Work}

A common way to improve the performance of low-resource ASR models is to leverage data from other high-resource settings.  Transfer learning and multitask learning are commonly used methods. Transfer learning \cite{DBLP:conf/rep4nlp/KunzeKKKJS17, DBLP:conf/interspeech/HuangYLG14a, DBLP:conf/interspeech/JoshiZMKL20}  first trains the model on the high-resource setting and then fine-tune it on the target data-scarce setting. The parameters learned from the first setting serve as a starting point, also known as supervised pre-training. In multitask learning \cite{huang2013cross, DBLP:conf/interspeech/KnillGRR14, chen2015multitask}, the model is simultaneously trained on multiple languages with shared components.  Both methods depend on labeled data from multiple languages to yield consistent improvements while large amount of unlabeled data cannot be used. 

Recently, self-supervised learning has received great attention as it does not require any labeled data.  Based on the training objectives, self-supervised methods can be categorized into reconstructive learning (recreating audio frames) and contrastive learning (discriminating true sample from set of negative samples). \citet{DBLP:journals/taslp/ChenHLWS19} use an autoencoder to perform full reconstruction. \cite{DBLP:journals/taslp/ChorowskiWBO19} use a high capacity WaveNet autoencoders to learn meaningful speech representations. They compare three different variants of constraints: a simple bottleneck, a Gaussian Variational Autoencoder (VAE) and a Vector Quantized VQE (VQ-VAE). Autoregressive predictive coding (APC) \cite{DBLP:conf/interspeech/ChungHTG19} reconstructs the future frame with an unidirectional encoder for phone classification and SR task. Masked reconstruction \cite{DBLP:conf/icassp/LiuYCHL20,DBLP:conf/icassp/LingLSK20,ling2020bertphone} has also been widely investigated which masks part of the input and learn to reconstruct it. In the research line of contrastive learning, CPC \cite{oord2018representation} uses an autoregressive model to classify future frames from negative examples. Wav2vec \cite{DBLP:conf/interspeech/SchneiderBCA19} evaluates the effectiveness of contrastive learning on speech recognition task.  \citet{DBLP:conf/emnlp/KawakamiWDBO20} and \citet{DBLP:conf/icassp/RiviereJMD20} show bi-directional and modified CPC transfers well across domains and languages.  Vq-wav2vec \cite{DBLP:conf/iclr/BaevskiSA20} uses a vector quantization module to learn discrete representations. They further introduce Wav2vec2.0 \cite{DBLP:conf/nips/BaevskiZMA20}, which masks the speech input in the latent space and solves a contrastive task defined over contextual representations in the masked region and a quantization of the latent representations. They show the discrete latent speech representations learnt by quantizer are related to phonemes. \citet{DBLP:journals/corr/abs-2006-13979} try the idea on multilingual settings, named XLSR. 

Some other recent work employ multitask learning strategy for speech representation learning. \cite{DBLP:conf/interspeech/PascualRSBB19} and \cite{DBLP:conf/icassp/RavanelliZPSMTB20} propose to learn a problem-agnostic speech encoder which jointly solves different self-supervised tasks, including reconstructive loss and contrastive loss. Compared to our work, they use only unlabeled data and the models are evaluated on speaker identification, emotion classification and ASR tasks. While we focus on improving ASR performance in low-resource scenarios with all availabel data.  \cite{DBLP:journals/corr/abs-2011-00093} alternatively minimize the unsupervised masked CPC loss and the supervised CTC loss. They focus on simplify the training pipeline and reports equivalent word error rate as in wav2vec2.0. In contrast, our method can improve the recognition performance significantly.

\section{UniSpeech}
\subsection{Problem Formulation}
Suppose we have $I$ datasets from high-resource settings, denoted as $\mathbb{L}= \{(\mathbf{X}_1^L,\mathbf{Y}_1^L), ..., (\mathbf{X}_I^L,\mathbf{Y}_I^L)\}$. Each dataset contains a large scale of audio-text pairs $(\bm{x}, \bm{y})$. We also have $J$ low-resource languages/domains for evaluation. For each, we have a large unlabeled dataset and a small labeled dataset, denoted as $\mathbb{M} = \{(\mathbf{X}_1^M), ..., (\mathbf{X}_J^M\}$ and $\mathbb{N} = \{(\mathbf{X}_1^N,\mathbf{Y}_1^N), ..., (\mathbf{X}_J^N,\mathbf{Y}_J^N)\}$ respectively.  Our goal is to leverage accessible large datasets $\mathbb{L}$ and $\mathbb{M}$ to learn robust representations and then fine-tune the model on $\mathbb{N}$ to improve the ASR performance on the low-resource settings. 
   
\subsection{Model Structure} \label{structure}

Our model architecture is shown in Figure \ref{model}. Following the design choices in wav2vec2.0 \cite{DBLP:conf/nips/BaevskiZMA20}, the model contains a convolutional feature encoder, a Transformer context encoder and a vector quantizer. 
The convolutional feature encoder $\mathcal{X} \to \mathcal{Z}$ maps raw audio $\bm{x}$ to a latent space $\mathcal{Z}$. It is composed of seven blocks of temporal convolution followed by layer normalization and GELU activation layer. The temporal convolutions in each block have 512 channels with strides (5,2,2,2,2,2,2) and kernel widths (10,3,3,3,3,2,2), resulting in each $z_t$ represents about 25ms of audio strided by 20ms. Then the representations $z_1, ...z_T$ are fed into the Transformer \cite{DBLP:conf/nips/VaswaniSPUJGKP17,DBLP:conf/naacl/DevlinCLT19} network $\mathcal{Z} \to \mathcal{C}$ to output context representations $c_1, ..., c_T$.  The Transformer network is equipped  with a convolutional layer with kernal size 128 and 16 groups to replace absolute positional embedding. We evaluate on two different settings with the same feature encoder: Base with 12 Transformer blocks, model dimension 768, inner dimension 3072 and 8 attention heads; and Large with 24 Transformer blocks,  model dimension 1024, inner dimension 4096 and 16 attention heads  Acting as an information bottleneck on the latent representation, the quantizer module $\mathcal{Z} \to \mathcal{Q}$ discretizes  $z_i$ to a finite set of speech representations $q_i$. The quantizer has $G=2$ codebooks with $V=320$ entries each. For each frame, we select two entries from two codebooks independently to obtain quantized representation $q$, resulting in over 100K codewords in total ($320^2$).

\subsection{Multitask Learning with Unified Representation}
Reconsidering the problem, we would like to learn a representation with the following two features: 1) Each frame's representation corresponds to a meaningful phonetic unit. 2) The representation is easy to adapt to the target domain SR task. To achieve this, we propose a multitask learning method with unified representation. In the pre-training stage, we jointly train the model on high-resource labeled dataset $\mathbb{L}$ and
low-resource dataset $\mathbb{M}$. Our training objectives are three parts: 1) Phonetic CTC loss $\mathcal{L}_{ctc}$ on dataset $\mathbb{L}$. It aligns the contextual representations with phonetic units. 2) Contrastive loss $\mathcal{L}_{c}$ on dataset $\mathbb{L}$. The loss closes the distance between the representations $c$ and the discrete features $q$, resulting in phonetically aware codebooks. 3) Contrastive loss on dataset $\mathbb{M}$. It adapts the model on the target language or domain.

Specifically, given a data  pair $(\bm{x},\bm{y})$, the model learns its context representations $c_1, ..., c_T$. We add a linear layer with softmax to predict a distribution over observed labels, including phoneme tokens and a blank token, denoted as  $p(\bm{\pi}|c_1, ..., c_T)$. A legal CTC path $\bm{\pi}$ is a variation of the transcription $\bm{y}$ by allowing occurrences of blank tokens and repetitions. The CTC objective trains the model to maximize the sum of  conditional probability of all possible legal paths:

\begin{equation} \label{ctc}
\small
    \mathcal{L}_{ctc} = - \text{log} p_{ctc}(\bm{y}|\bm{x}) = \sum_{\bm{\pi} \in \Phi_{\bm{x},\bm{y}}} p(\bm{\pi}|c_1, ..., c_T)
\end{equation} where $\Phi_{\bm{x},\bm{y}}$ is the set of all valid alignments. 

Through CTC supervised learning, the model can map each frame's representation $c_t$ to a phonetic unit explicitly. However, the learnt representation is located in the source domain and it is hard to be transferred to the target scenario with only limited labeled data. In order to generalize this model, we leverage the self-supervised contrastive learning on both labeled source data and unlabeled target data, i.e. $\{\mathbf{X}_i | (\mathbf{X}_i, \mathbf{Y}_i) \in \mathbb{L} \lor \mathbf{X}_i \in \mathbb{M}\}$.  

Given $\bm{x}$, we can obtain the feature representation $z_1 ... z_T$ with the convolutional feature encoder. During training, we mask some frames and fed the masked features $\tilde{\bm{z}}$ into the Transformer. We use the same mask strategy as in wav2vec2.0 that randomly sampling start indices with probability $p$ and mask the consecutive ten time steps.  The model uses quantizer output $\bm{q}$ as the contrastive targets, while input to the quantizer is unmasked. First, $z_i$ is mapped to $l \in \mathbb{R}^{G\times V}$ logits, where $G$ is the number of codebooks and $V$ is the number of entries for each. Then the module relies on Gumbel softmax \cite{jang2016categorical} to choose one discrete entry $e$ from each codebook based on probability 
\begin{equation}
\small
    p_{g,v} = \frac{\text{exp}(l_{g,v}+n_v)/\tau}{\sum_{k=1}^V\text{exp}(l_{g,v}+n_v)/\tau},
\end{equation} where $\tau$ is a non-negative temperature, $n=-\text{log}(-\text{log}(u))$ and $u$ are uniform samples from $\mathcal{U}(0, 1)$. In the forward pass, the quantizer finds a nearest prototype to the input $z$ from each codebook, denoted as $e_g(i)$, where  $i = \text{argmax}_jp_{g,j}$. We concatenate the resulting vectors $e_1,..., e_G$ and apply a linear transformation to obtain $q$. $q$ has the same dimension as Transformer encoder.  In the backward pass, the gradient of the loss with respect to the pre-quantized vector $z$ is approximated using the straight-through estimator, that is $\frac{\partial\mathcal{L}}{\partial z} \approx \frac{\partial \mathcal{L}}{\partial q}$.

For each $c_t$ centered over masked time step $t$, the model needs to identify the true quantized latent speech representation $q_t$ in a set of $K+1$ quantized candidates $\rm{Q}_t$. The $K$ distractors are uniformly sampled from the other timesteps from the same utterance. This frees up the model from using its capacity to represent speaker-dependent information and instead focuses on phonetic features.  The loss is defined as 
\begin{equation}\small
    \mathcal{L}_{c} = - \text{log}\frac{\text{exp}(sim(c_t, q_t)/\kappa)}{\sum_{\tilde{q} \sim \rm{Q}_t}\text{exp}(sim(c_t, \tilde{q})/\kappa)}
\end{equation} where we use cosine similarity $sim(a, b) =\frac{a^Tb}{\| a \| \| b \|} $. The contrastive loss encourages the quantizer to produce vectors which lie close to the contextual representations $c$. As we train the objective on the joint set of $\mathbb{L}$ and $\mathbb{M}$, the codebook can generalize  at both source and target domain.

The objective is augmented by a codebook diversity loss with a loss weight 0.1. It encourages the equal use of all entries by maximizing the entropy of the averaged softmax distribution $l$ over the codebook entries:
\begin{align} \label{self_loss} \small
  &\mathcal{L}_{self}  = \mathcal{L}_c + 0.1 * \mathcal{L}_d \\
   \text{where\,\,\,\,} & \mathcal{L}_{d}  = \frac{1}{GV}\sum_{g=1}^{G}\sum_{v=1}^{V}\bar{p}_{g,v}\text{log}\bar{p}_{g,v}
\end{align}
The final pre-training loss can be defined as:

\begin{equation} \label{multi_task}
\small
    \mathcal{L} =  \sum_{(\bm{x},\bm{y}) \in \mathbb{L}}(\alpha \mathcal{L}_{ctc} + (1-\alpha) \mathcal{L}_{self}) + \sum_{(\bm{x}) \in \mathbb{M}} \mathcal{L}_{self} 
\end{equation} where $\alpha$ is the weight for loss combination on dataset $\mathbb{L}$.

The quantization module was qualitatively shown to learn a representation which separates phonetic content within an utterance from the speaker identity \cite{DBLP:conf/nips/BaevskiZMA20}. Moreover they discover the tokens learnt in an unsupervised manner can be mapped to phonemes in a limited setting. However, it cannot guarantee that the discrete representations are as useful as those learnt by the supervised learning for the ASR tasks. And we find the above multitask method leads to limited gain (as shown in the Table \ref{one2one} of the experiment part).
To address it, when calculating the CTC loss, we replace the continuous representation $c$ with its quantized versions $q$ with probability $r$.  Mathematically, the conditional probability of Eq. \ref{ctc} is changed as 
\begin{equation} \label{UniSpeech_loss}
\small
   \sum_{\bm{\pi} \in \Phi_{\bm{x},\bm{y}}}\prod_{t=1}^T p(\bm{\pi}|c_1 ... c_T) \longrightarrow \sum_{\bm{\pi} \in \Phi_{\bm{x},\bm{y}}} \prod_{t=1}^T p(\bm{\pi}|c'_1 ... c'_T)
\end{equation} where $c'_i$ is either $c_i$ or $q_i$. Since $\bm{y}$ is a phoneme sequence, predicting $\bm{y}$ with $q_i$ can explicitly guide the quantizer to cluster phonemes and learn SR specific knowledge into codebooks.

With Eq. \ref{UniSpeech_loss}, representations in supervised learning and unsupervised learning are forced to project into the same space, and it avoids the two objective functions optimize themselves individually. Although the method is simple, it is effective according to our experiment results.  Since the representations are unified in two tasks and different languages, we call our model UniSpeech. 

After pre-training, we fine-tune the model with $\mathbb{N}$ using the CTC loss. During finetuning, we replace the pretrained CTC layer with a new layer to represent the target vocabulary. We freeze the weights of the feature encoder and only fine-tune the Transformer part. It should be noted that we can use advanced SR objective in supervised learning such as transducer loss \cite{Graves-RNNSeqTransduction} or seq2seq cross-entropy loss \cite{chan2016listen}. Our future work will explore it.

\section{Experiments}
We evaluate our methods on Multilingual ASR task and domain transfer ASR task. 
\subsection{Multilingual ASR}
\subsubsection{Setup}
\paragraph{Dataset} Regarding multilingual ASR, we first train the UniSpeech model on high-resource languages, and then transfer it to low-resource languages. We employ the CommonVoice (CV) dataset \cite{ardila2019common} \footnote{\url{https://commonvoice.mozilla.org/en/datasets}. We use the June 2020 release version for training our models.}, which is a multilingual corpus of read speech comprising more than 5k hours of speech data in 60 languages.  To be comparable with XLSR \cite{DBLP:journals/corr/abs-2006-13979}, we consider the following eight languages for evaluation\footnote{Files in the test set of Turkish(tr) and Chinese(zh) are missing in CommonVoice June2020 release, so we exclude these languages.}: \textit{Spanish (es), French (fr), Italian (it), Kyrgyz (ky), Dutch (nl), Russian (ru), Swedish (sv)} and \textit{Tatar (tt)}. \textit{English (en)} is always regarded as a high-resource language. Our pre-training data is not exactly same as  \citet{DBLP:journals/corr/abs-2006-13979} as we use different dataset version, but we use the same data size for each language without selection.  We define three settings based on the number of pre-training languages and fine-tuning langauges: \textit{one-to-one}, \textit{many-to-one} and \textit{many-to-many}. The  pre-training details will be illustrated in the corresponding experiment settings. For fine-tuning, we use the evaluation splits from \citet{DBLP:conf/icassp/RiviereJMD20}, which contains 1 hour paired data for training, 20 minutes for validation and 1 hour for testing.   We retrieve phoneme transcriptions by running open-source phonemizer\footnote{\url{https://github.com/bootphon/phonemizer}} and report PER following prior work \cite{DBLP:journals/corr/abs-2006-13979}. 

\paragraph{Implementation Details} Models are implemented in fairseq \cite{ott2019fairseq}. To train the UniSpeech model, we use mask probability $p = 0.05$, loss weight $\alpha = 0.5$ and replace probability $r = 0.5$ unless otherwise stated. During pre-training, we crop each utterance to 250k samples for Base model and 320k samples for Large model. Each batch on one GPU contains max up to 1.4m samples for Base and 1.2m samples for Large. The models are trained on 64 GPUs. We use Adam optimizer where the learning rate is warmed up for the first 10\% of updates to a peak of 5e-4(Base) or 1e-3(Large) and then linearly decayed over a total of 250k updates.  The model is fine-tuned with 2 GPUs. We still use Adam optimizer and the learning rate is warmed up for 2k updates to 2e-5, keep constant for 8k updates and then linearly decay for 10k updates. Dropout 0.1 is always used for both pre-training and finetuning. 

\begin{table*}[t]
\vskip 0.15in
\begin{center}
\begin{small}
\begin{tabular}{l|c|c | llllllll|l}
\toprule
Model    & \multicolumn{2}{c|}{pre-trained data}   & es  & fr   & it   & ky   & nl   & ru   & sv   & tt   & avg  \\ \cline{2-3}
  & $\mathbb{L}$ & $\mathbb{M}$ &  &  & & & & & & &  \\ \midrule
\multicolumn{3}{l|}{Number of unlabeled data (\#$\rm{CV_{mo}}$)}   &  168h  & 353h  & 90h & 17h & 29h & 55h&  3h  & 17h  &  \\ \midrule\midrule
\multicolumn{11}{l}{\textit{Baseline results that copy from previous literature }}  \\ \hline

m-CPC \cite{DBLP:conf/icassp/RiviereJMD20} &  & $\rm{LS_{100h}}$   & 38.7  & 49.3 &  42.1 & 40.7  &  44.4  & 45.2  &  48.8  & 44.0  &  44.2 \\  
m-CPC \cite{DBLP:conf/icassp/RiviereJMD20}  &  & $\rm{LS_{360h}}$  & 38.0 & 47.1 & 40.5 & 41.2 & 42.5 & 43.7 & 47.5  & 42.0  & 42.8 \\

XLSR-English \cite{DBLP:journals/corr/abs-2006-13979} &  & $\rm{CV_{en}}$*  & 13.7 & 20.0 & 19.1 & 13.2 & 19.4 & 18.6 & 21.1 & 11.5 & 17.1 \\
$\text{XLSR-Mono}^{+}$ \cite{DBLP:journals/corr/abs-2006-13979} & &  $\rm{CV_{mo}}$* & 6.8 & 10.4 & 10.9 & 29.6 & 37.4 & 11.6 & 63.6 & 21.4 & 24.0 \\ 

\hline
\multicolumn{11}{l}{\textit{Re-run baselines}}  \\ \hline
CTC-Transfer  & $\rm{CV_{en}}$  & & 12.6 & 16.7 & 16.4 & 12.9 & 17.5 & 17.5 & 20.7  & 11.2 & 15.7 \\ 
XLSR &  & $\rm{CV_{en}}$& 12.4 & 16.9 & 17.2 & 13.7 & 18.2 & 19.0 & 22.4 & 11.4 & 16.4 \\ 
$\text{XLSR}^{+}$  &  & $\rm{CV_{en, mo}}$ & 6.5 & 9.0 & 9.3 & 8.2 & 9.8 & 10.1 & 20.5 & 7.4 &  10.1  \\
XLSR-L  & & $\rm{CV_{en}}$ & 11.0 & 14.8 & 15.6 & 11.3 & 16.2 & 17.0 & 19.6 & 10.7 & 14.7 \\
$\text{XLSR-L}^{+}$  &  & $\rm{CV_{en, mo}}$ & 5.9 & 7.9 & 8.5  & 7.9 & 9.4 & 9.6  & 20.2 & 7.0 &  9.6 \\
\hline
\multicolumn{11}{l}{\textit{Our models}}  \\ \hline
UniSpeech    & $\rm{CV_{en}}$  &   & 10.9 & 14.8 & 15.2 & 11.4 & 16.2 & 16.1 & 19.3 & 9.6 & 14.2 \\
\,\,\,\,\,\,$r$=0  & $\rm{CV_{en}}$   &   & 12.1 & 16.5 & 16.3 & 12.3 & 17.2 & 16.8 & 20.5 & 10.9 & 15.3 \\
UniSpeech$^{+}$     &   $\rm{CV_{en}}$    &   $\rm{CV_{mo}}$   & {5.7} & {7.9} & {8.1} & {6.8} & {9.3}  &  {8.6}  & {17.7} & {6.0} & {8.8} \\
UniSpeech-L & $\rm{CV_{en}}$  &   & 10.2 & 13.3 & 14.6 & 10.8 & 15.3 & 16.0 & 19.3 & 9.6 & 13.6 \\
UniSpeech-L$^{+}$      &   $\rm{CV_{en}}$    &   $\rm{CV_{mo}}$   & {4.7} & 6.2 & 6.8 & {6.1} & 6.8  &  {8.3}  & {17.1} & 5.5 & 7.7 \\
\bottomrule
\end{tabular}
\caption{One-to-one evaluation results. The numbers listed in the table are phone error rate. The last column is the averaged PER on eight languages. *: They use different version of the CommonVoice dataset, but the data size is the same as ours. +: The unlabeled data of target languages are used in pre-training stage. -L: Large model. \label{one2one}}
\end{small}
\end{center}
\end{table*}

\subsubsection{Results}
In our experiments, we mainly compare our method with two baselines: 1) CTC-Transfer, which pre-trains a CTC model on high-resource dataset $\mathbb{L}$ and then fine-tunes it on low-resource dataset $\mathbb{N}$;  and 2) XLSR \cite{DBLP:journals/corr/abs-2006-13979}, which is an unsupervised learning method based on wav2vec2.0. As their pre-training dataset is unavailable, we re-run the XLSR experiments on our datasets. In addition, we also list results from literature which use the same finetune dataset and test dataset. We mainly report results for Base setting unless otherwise stated.

\paragraph{One-to-one}
In this setting, we pre-train the model on English dataset and transfer it to single low-resource languages. We assume a 1350 hours of labeled English dataset (i.e. $\mathbb{L} = \rm{CV_{en}}$) and an unlabeled dataset (i.e. $\mathbb{M} = \rm{CV_{mo}}$) for each low-resource languages are accessible. The data size for each $\rm{CV_{mo}}$ is listed in Table \ref{one2one}.  We train a CTC transfer baseline with only supervised data $\mathbb{L} = \rm{CV_{en}}$ as well as two unsupervised baselines, XLSR with $\mathbb{M} = \rm{CV_{en}}$ and $\text{XLSR}^+$ with $\mathbb{M} = \{\rm{CV_{en}}, \rm{CV_{mo}}\}$. In following parts, we use superscript $+$ to denote that the target language unlabeled dataset $\mathcal{M}$ has been used in the pre-training stage. For UniSpeech, we also show two results: one is pre-trained on only $\rm{CV_{en}}$ to be comparable to CTC-Transfer baseline, another is pre-trained on $\rm{CV_{en}}$ and $\rm{CV_{mo}}$, denoted as $\text{UniSpeech}^{+}$.  After pre-training, all the models are fine-tuned on the 1 hour labeled low-resouce dataset $\mathbb{N}$.  We also list results from previous work: \citet{DBLP:conf/icassp/RiviereJMD20} use modified CPC model to pre-train phoneme representation on either 100  or 360 hours of Librispeech clean data and transfer it to CommonVoice low-resource languages, denoted as m-CPC. \citet{DBLP:journals/corr/abs-2006-13979} train XLSR-English with dataset $\rm{CV_{en}}$ and $\text{XLSR-Mono}^+$ with only $\rm{CV_{mo}}$. 

From Table \ref{one2one}, we can see that CTC-Transfer is better than XLSR as it uses the label information in English data. However, when the target language unpaired data is available, the performance of $\text{XLSR}^+$ outperforms CTC-Transfer significantly, indicating in-language data is the key to the success of unsupervised method. Compared with CTC-Transfer and XLSR, our UniSpeech obtain PER reductions of  9.6$\%$ and 13.4$\%$ respectively, which is mainly because our method combines transfer learning and self-supervised learning, and the two methods are complementary. However, when we set the replace probability $r$ as 0, our method degrades to multitask learning and  it is worse than UniSpeech by 7.7$\%$ relatively. Furthermore, when target unlabeled data is available, $\text{UniSpeech}^+$ achieves 8.8 PER score, outperforming $\text{XLSR}^{+}$ by relative 12.9\%. This indicates our model can well utilize both supervised and unsupervised data and it learns robust speech representation which is easily transferred across different languages. The conclusion is the same for Large setting. The best result for the one-to-one setting is obtained by UniSpeech-L$^+$, which gets 7.7 PER on average, a relative PER reduction of 19.8\% compared to XLSR$^+$ Large model.

\begin{table}[t]
\begin{center}
\begin{small}
\begin{tabular}{l|llllll} 
\toprule
Method                          & ky   & nl   & ru   & sv   & tt   & avg  \\ \midrule
\multicolumn{7}{l}{\textit{Baselines from \citet{DBLP:journals/corr/abs-2006-13979}}}  \\ \hline
$\text{XLSR-10}^{+}$    &   8.4  &  16.1  &  11.0  &  20.7  &  7.6  &   12.8 \\
$\text{XLSR-10-L}^{+}$  & 7.0  & 14.0  & 9.3  & 20.6  & 7.2  &  11.6 \\ \midrule
\multicolumn{7}{l}{\textit{Re-run baselines}}  \\ \midrule
CTC-Transfer       & 12.2 & 18.2 & 17.0 & 20.3 & 10.8 & 15.7 \\
XLSR                     & 10.9 & 16.1 & 15.5 & 19.6 & 9.2  & 14.3 \\
$\text{XLSR}^{+}$  &  7.2 & 13.2 & 11.0 & 19.2 & 6.5 &  11.4    \\ 
XLSR-L  & 10.4 & 15.0 & 15.6 & 18.7 & 9.2 & 13.8 \\
$\text{XLSR-L}^{+}$  & 8.5  & 11.0  & 9.6 & 18.7 & 6.1  &   10.8  \\

 \midrule
\multicolumn{7}{l}{\textit{Our models}} \\ \midrule
UniSpeech  & 10.1 & 14.3 & 14.2 & 17.3 & 8.6  & 12.9 \\
\,\,\,\,\,\,$r$=0 & 11.5 & 16.3 & 15.5  & 19.2 & 10.0 & 14.5 \\
$\text{UniSpeech}^{+}$   &   6.2   &   9.0   &  8.2   &  15.9    &  5.6  & 9.0     \\
UniSpeech-L & 9.2 & 13.5 & 13.9 & 17.1 & 8.5  & 12.4 \\
$\text{UniSpeech-L}^{+}$   &  5.3  & 6.1  & 7.5  & 16.0  &  4.8  &  7.9   \\
\bottomrule

\end{tabular}\caption{Many-to-one evaluation results.  +: The unlabeled data of target languages are used in pre-training stage. -L: Large model. \label{many2one}}
\end{small}
\end{center}

\end{table}
\begin{table}[t]
\begin{center}
\begin{small}
\begin{tabular}{l|llllll}
\toprule
Method                          & ky   & nl   & ru   & sv   & tt   & avg  \\ \midrule
\multicolumn{7}{l}{\textit{Baselines from \citet{DBLP:journals/corr/abs-2006-13979} (share vocabulary)}}  \\ \hline
$\text{XLSR-10}^{+}$  & 8.8    &  16.5  &  11.6  &  21.4  &  8.7  & 13.4   \\ \midrule
\multicolumn{7}{l}{\textit{Re-run baselines (share vocabulary) }}  \\ \midrule
CTC-Transfer   &  13.2  & 18.3  &  17.7  &  21.6  &  11.0  &  16.4  \\
XLSR                     & 12.0 & 17.9 & 17.2 & 21.2 & 10.5 & 15.8 \\
$\text{XLSR}^{+}$   &   6.6   &    13.1  & 10.6     & 21.6     & 6.3   & 11.6 \\
 \midrule
\multicolumn{7}{l}{\textit{Our models (share vocabulary) }} \\ \midrule
UniSpeech    &  11.9  & 16.5    &  15.9    & 20.5  & 9.9  & 14.9   \\
$\text{UniSpeech}^{+}$ &  {6.5}   &  {11.9}    & {9.9} & {19.5}  &  {5.9} & {10.7}  \\
\midrule \midrule
\multicolumn{7}{l}{\textit{Baselines from \citet{DBLP:journals/corr/abs-2006-13979} (separate vocabulary) }}  \\ \hline
$\text{XLSR-10}^{+}$  & 8.6  & 16.3  &  11.2  & 21.0  & 8.3  &  13.1 \\
\multicolumn{7}{l}{\textit{Re-run baselines (separate vocabulary) }}  \\ \midrule
CTC-Transfer  &  12.8    &  18.3   &  17.8  &   21.6  &  11.2   &   16.3 \\
XLSR   & 11.7  &  17.6  & 16.6  & 21,1  & 10.1  & 15.4  \\
$\text{XLSR}^{+}$   & 6.8  & 13.3  & 10.6  & 21.2  & 6.5  & 11.7 \\
 \midrule
\multicolumn{7}{l}{\textit{Our models (separate vocabulary) }} \\ \midrule
UniSpeech   & 11.3  & 16.2  &  15.6    &   20.0  & 9.9    & 14.6     \\
$\text{UniSpeech}^{+}$ & {7.0} & {12.2} &  {9.9} &  {19.5} & {6.7} &  {11.0} \\ \bottomrule
\end{tabular} \caption{Many-to-many evaluation results.  +: The unlabeled data of target languages are used in pre-training stage. L: Large model. \label{many2many}}
\end{small}
\end{center}
\end{table}

\paragraph{Many-to-one}
We further demonstrate the effectiveness of cross-lingual transfer on low-resource languages. In this setting, we use 4 labeled datasets for pre-training, including 1350h \textit{en}, 168h \textit{es}, 353h \textit{fr}, and 90h \textit{it} (i.e. $\mathbb{L} = \{\rm{CV_{en}},\rm{CV_{es}},\rm{CV_{fr}},\rm{CV_{it}}\}$). During pre-training, we form multilingual batches by sampling speech utterance from a multinomial distribution $(p_l)_{l\in\mathbb{L}}$ where $p\sim(\frac{n_l}{N})^{0.5}$, $n_l$ being the number of pre-training hours of language $l$ and $N$ being the total number of hours. The performances are evaluated on the other five languages.  The monolingual unlabeled data for each target language (i.e. $\mathbb{M} = \rm{CV_{mo}}$) is also used in experiments $\text{XLSR}^{+}$ and $\text{UniSpeech}^{+}$. The baselines from \citet{DBLP:journals/corr/abs-2006-13979}  are denoted as $\text{XLSR-10}^{+}$ and $\text{XLSR-10-L}^{+}$. They use 1350 hours of unlabeled data from 10 languages and the target language data is always used in their pre-training stage.

In Table \ref{many2one}, the observation is consistent with one-to-one setting. The UniSpeech reaches relative 17.8\% and 9.8$\%$ PER reduction compared to CTC-Transfer and XLSR respectively, indicating that our method is effective on the cross-lingual transfer task. An interesting phenomenon is the performances of CTC-Transfer and even the UniSpeech with $r=0$ are worse than XLSR, which is different from the conclusion in the previous setting.  A possible explanation is that it is hard to deal with multiple vocabularies in CTC pre-training. As there are many overlapped phonemes appear in multiple vocabularies, some of them share similar pronunciation while the others have totally different pronunciations. Instead, the unsupervised method does not suffer from this issue. This indicates the codebook in unsupervised learning has the ability to adapt to multiple languages.  When the target unlabeled data is available, $\text{UniSpeech}^{+}$ outperforms $\text{XLSR}^{+}$ by 21.1\% relative PER.  It also significantly outperforms $\text{XLSR-10}^{+}$ which uses unpaired data from 10 languages.  For Large model, our UniSpeech$^{+}$ even improves baseline by 26.9\% relative. It shows UniSpeech is more robust and transferable than either supervised learning or unsupervised learning alone. Furthermore, many-to-one results are better than one-to-one results on the 5 unseen low-resource languages. It suggests that large and diverse training data is beneficial to our model.

\paragraph{Many-to-Many}
We also evaluate the model for multilingual fine-tuning. In the pre-training stage, we use the same pre-trained model as in the many-to-one setting for UniSpeech experiment. For $\text{UniSpeech}^+$, instead of pre-training one model for each low-resource scenario, we merge the 5 unlabeled low-resource datasets and pre-train the model on the joint set. In the fine-tuning stage, the 5 labeled datasets are always merged. As there are overlapped phonemes across languages, we can either regard each as a single label for shared vocabulary or as different labels with different language ids for separate vocabulary.  

As Table \ref{many2many} shows, UniSpeech outperforms CTC-Transfer and XLSR by 10.4$\%$ and 5.1$\%$ respectively. When the unlabeled datasets from target languages are used, the gain for UniSpeech$^+$ against XLSR$^+$  is 6$\%$ PERR. 
 The overall PER is higher than the many-to-one setting, because the multi-lingual outputs make the task harder. The performances of shared vocabulary and separated vocabulary are similar. After checking the shared phoneme vocabulary, we find that the same phoneme unit generated by the phonemizer may have different pronunciation and thus represent different sounds in different languages. A better universal phonemizer is worth trying in the future.  
 \begin{figure*}
    \centering
    \includegraphics[width=0.8 \textwidth]{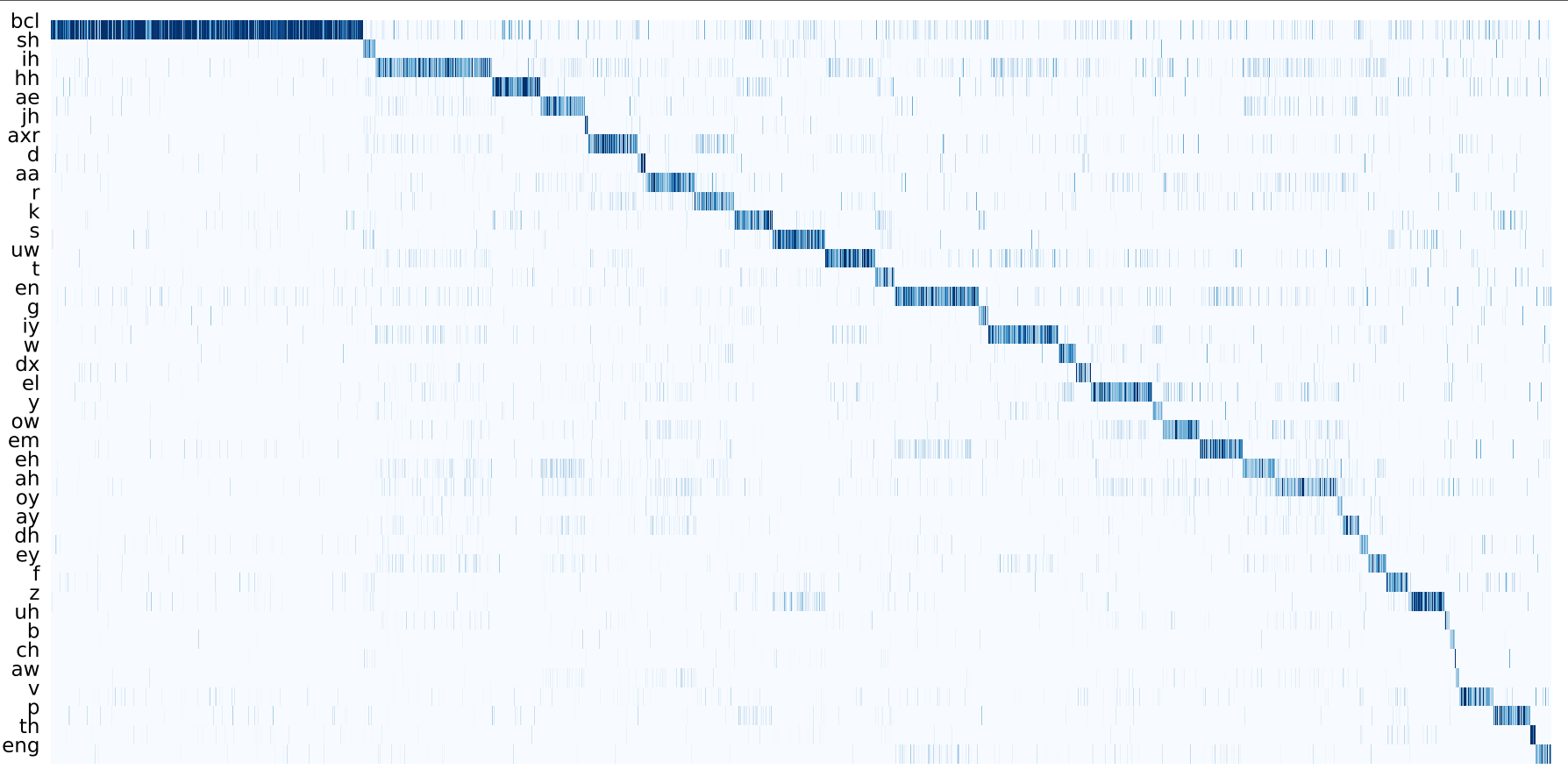}
  \caption{Visualization of the co-occurrence between discrete latent speech representations and phonemes.  }
  \label{hotmap}
\end{figure*}
 \begin{table}[!t]
\centering
\begin{small}

\begin{tabular}{l|cc}
\toprule
           &  \multicolumn{2}{c}{Tedlium3} \\
           & dev          & test          \\ \midrule
\multicolumn{3}{l}{\textit{Baselines from \citet{DBLP:conf/emnlp/KawakamiWDBO20}}}  \\ \midrule
LogFilterbank   &  18.75  &  19.31  \\
CPC-Librispeech   &  15.28  & 15.87  \\
CPC-8k   & 13.67  & 13.88 \\ \midrule
\multicolumn{3}{l}{\textit{Re-run baselines and our Model}}  \\ \midrule
CTC-Transfer &  8.3  & 8.0        \\
Wav2vec2.0  & 8.3    & 8.1           \\
UniSpeech   & 7.6   & 7.6           \\
\bottomrule

\end{tabular} \caption{Domain transfer results to test the representation robustness on domain shifting.} 
\label{domain_transfer}
\end{small}
\vspace{-4mm}
\end{table}

\subsection{Domain Transfer}
We also conduct an experiment for domain transfer task. We use our UniSpeech model to pre-train phoneme representations in reading English domain, namely on Librispeech, and transfer them to spoken English domain on Tedlium3  \cite{hernandez2018ted} dataset. We only use the 960 hours of labeled speech as pre-training dataset in this experiment. To train the UniSpeech model, we also extract the phoneme sequences by phonemizer to calculate the CTC loss. During fine-tuning, we discard the phoneme CTC layer and use character based CTC loss. We reports WER on dev and test sets. The model is pre-trained on 64 GPUs to 400k steps and fine-tuned on 8 GPUs to 320k steps. Other training parameters are the same in the multilingual experiments.

We compare our model with supervised CTC-Transfer and unsupervised wav2vec2.0 pre-training. As we use character-level fine-tuning loss, we report results using  character-level pre-training for CTC-Transfer baseline. This leads to better performance than phonetic-level pre-training. We also compare with results from \citet{DBLP:conf/emnlp/KawakamiWDBO20}. They first train a bidirectional CPC model on an unlabeled audio dataset, either Librispeech or 8k mixed audio dataset spanning a range of recording conditions, noise levels, speaking styles and languages. Next they freeze the model's parameter and use its output representations as input to train a TDNN based CTC model. Their methods are denoted as CPC-Librispeech and CPC-8k respectively. They also report results when using LogFilterbank as input feature.

From Table \ref{domain_transfer}, we can draw the following conclusions. There is no doubt that wav2vec 2.0 outperforms CPC-based models largely because of a better network structure (Transformer v.s. TDNN) and a better pretraining method. The wav2vec2.0 baseline obtains similar performance compared with CTC transfer learning,  indicating that self-supervised learning can learn powerful speech representation. UniSpeech is better than wav2vec 2.0 and CTC transfer, which shows the advantages of our model on domain transfer task.

\subsection{Discussion and Analysis}
\paragraph{Analysis of Discrete Representations }
In this section, we investigate whether the discrete latent speech representations $q$ learnt by the quantizer can be mapped to the meaningful phonetic units. Following \citet{DBLP:conf/nips/BaevskiZMA20}, we compute the discrete latents on the training data of TIMIT, which contains 5 hours of audio recordings with human annotated phonemes. We use the multilingual UniSpeech model without any fine-tuning.  We then compute the conditional probability $p(\text{phn}|q_t)$ based on the co-occurrence between phonemes and the latents. The alignments  are built by choosing the phoneme which is most represented in the receptive field of each $q$. Figure \ref{hotmap} shows that many discrete latents appear to specialize in specific phonetic sounds, indicating our methods can obtain a good alignment between latents and labeled phonemes. The silence phoneme (bcl) is aligned with the most latents, since there are many blanks tokens in CTC training and TIMIT data has slience in every utterance. 

We define two metrics to quantitatively evaluate our method against the baselines.  1) The number of active discrete codewords $|Q|$ calculated with TIMIT training data. There are over 100k entries by combining two codebooks and most of them are not active (not triggered by the gumbel softmax layer for any frames of TIMIT training data). The more active latents, the more diverse the codebook. 2) The average entropy of alignments, which is computed by $\frac{\sum_{q}^{|Q|}\sum_{phn} p(\text{phn}|q_t)\text{log}p(\text{phn}|q_t)}{|Q|}$. High entropy indicates $p(\text{phn}|q_t)$ is closed to a uniformed distribution, thus a low entropy is preferred. 
We compare our model to multilingual XLSR. For UniSpeech, the number of active codewords is 25743 and the entropy is 0.83. While for XLSR, the two numbers are 3645 and 1.34 respectively. This indicates our model learns a more diverse codebook and it's better at phoneme clustering. It is a possible interpretation of why our model outperforms XLSR on CommonVoice dataset with the same diversity loss weight. 

\paragraph{Hyperparameter impact} 
Table \ref{hyper_tune} shows the impact of various hyperparameter choices. The experiments are conducted in the one-to-one setting. First, we show the impact of different replacement probability $r$. When we set $r$ = 0, the UniSpeech model degrades to a simple multitask model and the performance drops about 7.8\% relatively. The model performs similarly when $r$ ranges from 0.3 to 0.7. We set $r$ as 0.5 in all experiments as it performs best according to our hyperparameter search. Setting the loss weight $\alpha$ too low leads to lower performance. While increasing it from 0.5 to 0.7 leads to no improvements. Finally, we find that increasing the mask probability has little impact on the model's performance.

\begin{table}[t]
\vskip 0.15in
\begin{center}
\begin{small}
\begin{tabular}{ll}
\toprule
Baseline ($r$ = 0.5, $\alpha$ = 0.5, $p$ = 0.05) & 14.2 \\ \midrule
\multicolumn{2}{l}{Replacement probability} \\
$r$ = 0  & 15.3 \\
$r$ = 0.3 & 14.6 \\
$r$ = 0.7 & 14.5 \\ \midrule
\multicolumn{2}{l}{Loss weight} \\ 
$\alpha$ = 0.3 & 15.0 \\
$\alpha$ = 0.7 & 14.3 \\ \midrule
\multicolumn{2}{l}{Mask probability} \\
$p$ = 0.065 & 14.2 \\
$p$ = 0.075 & 14.2 \\ \bottomrule

\end{tabular} 
\caption{The impact of different hyper-parameters. }
\label{hyper_tune}
\end{small}
\end{center}
\vskip -0.3in
\end{table}

\section{Conclusion}
In this paper,  we propose the UniSpeech to learn speech representations with unlabeled and labeled data, in which supervised CTC labeling and phonetically-aware contrastive learning  are unified with a multitask learning framework. Unispeech consists of a convolutional feature extractor, a Transformer encoder and a quantizer, and the quantizer is explicitly guided to learn SR-specific information. The results show that UniSpeech outperforms both self-supervised and supervised pre-training alone by a large margin on multilingual SR and domain transfer tasks. In the future, we will scale up our model with over one million hours of unlabeled data, explore more complex SR architecture as well as the usage of  pre-trained language model in speech recognition. 

\nocite{}

\bibliography{example_paper}

\begin{thebibliography}{41}
\providecommand{\natexlab}[1]{#1}
\providecommand{\url}[1]{\texttt{#1}}
\expandafter\ifx\csname urlstyle\endcsname\relax
  \providecommand{\doi}[1]{doi: #1}\else
  \providecommand{\doi}{doi: \begingroup \urlstyle{rm}\Url}\fi

\bibitem[Ardila et~al.(2019)Ardila, Branson, Davis, Henretty, Kohler, Meyer,
  Morais, Saunders, Tyers, and Weber]{ardila2019common}
Ardila, R., Branson, M., Davis, K., Henretty, M., Kohler, M., Meyer, J.,
  Morais, R., Saunders, L., Tyers, F.~M., and Weber, G.
\newblock Common voice: A massively-multilingual speech corpus.
\newblock \emph{arXiv preprint arXiv:1912.06670}, 2019.

\bibitem[Baevski et~al.(2020{\natexlab{a}})Baevski, Schneider, and
  Auli]{DBLP:conf/iclr/BaevskiSA20}
Baevski, A., Schneider, S., and Auli, M.
\newblock vq-wav2vec: Self-supervised learning of discrete speech
  representations.
\newblock In \emph{{ICLR} 2020}. OpenReview.net, 2020{\natexlab{a}}.

\bibitem[Baevski et~al.(2020{\natexlab{b}})Baevski, Zhou, Mohamed, and
  Auli]{DBLP:conf/nips/BaevskiZMA20}
Baevski, A., Zhou, Y., Mohamed, A., and Auli, M.
\newblock wav2vec 2.0: {A} framework for self-supervised learning of speech
  representations.
\newblock In Larochelle, H., Ranzato, M., Hadsell, R., Balcan, M., and Lin, H.
  (eds.), \emph{NeurIPS 2020}, 2020{\natexlab{b}}.

\bibitem[Chan et~al.(2016)Chan, Jaitly, Le, and Vinyals]{chan2016listen}
Chan, W., Jaitly, N., Le, Q., and Vinyals, O.
\newblock Listen, attend and spell: A neural network for large vocabulary
  conversational speech recognition.
\newblock In \emph{ICASSP}, pp.\  4960--4964, 2016.

\bibitem[Chen \& Mak(2015)Chen and Mak]{chen2015multitask}
Chen, D. and Mak, B. K.-W.
\newblock Multitask learning of deep neural networks for low-resource speech
  recognition.
\newblock \emph{IEEE/ACM Transactions on Audio, Speech, and Language
  Processing}, 23\penalty0 (7):\penalty0 1172--1183, 2015.

\bibitem[Chen et~al.(2019)Chen, Huang, Lee, Wang, and
  Shen]{DBLP:journals/taslp/ChenHLWS19}
Chen, Y., Huang, S., Lee, H., Wang, Y., and Shen, C.
\newblock Audio word2vec: Sequence-to-sequence autoencoding for unsupervised
  learning of audio segmentation and representation.
\newblock \emph{{IEEE} {ACM} Trans. Audio Speech Lang. Process.}, 27\penalty0
  (9):\penalty0 1481--1493, 2019.

\bibitem[Chiu et~al.(2018)Chiu, Sainath, Wu, Prabhavalkar, Nguyen, Chen,
  Kannan, Weiss, Rao, Gonina, Jaitly, Li, Chorowski, and
  Bacchiani]{DBLP:conf/icassp/ChiuSWPNCKWRGJL18}
Chiu, C., Sainath, T.~N., Wu, Y., Prabhavalkar, R., Nguyen, P., Chen, Z.,
  Kannan, A., Weiss, R.~J., Rao, K., Gonina, E., Jaitly, N., Li, B., Chorowski,
  J., and Bacchiani, M.
\newblock State-of-the-art speech recognition with sequence-to-sequence models.
\newblock In \emph{{ICASSP} 2018}, pp.\  4774--4778. {IEEE}, 2018.

\bibitem[Chorowski et~al.(2019)Chorowski, Weiss, Bengio, and van~den
  Oord]{DBLP:journals/taslp/ChorowskiWBO19}
Chorowski, J., Weiss, R.~J., Bengio, S., and van~den Oord, A.
\newblock Unsupervised speech representation learning using wavenet
  autoencoders.
\newblock \emph{{IEEE} {ACM} Trans. Audio Speech Lang. Process.}, 27\penalty0
  (12):\penalty0 2041--2053, 2019.
\newblock \doi{10.1109/TASLP.2019.2938863}.
\newblock URL \url{https://doi.org/10.1109/TASLP.2019.2938863}.

\bibitem[Chung et~al.(2019)Chung, Hsu, Tang, and
  Glass]{DBLP:conf/interspeech/ChungHTG19}
Chung, Y., Hsu, W., Tang, H., and Glass, J.~R.
\newblock An unsupervised autoregressive model for speech representation
  learning.
\newblock In Kubin, G. and Kacic, Z. (eds.), \emph{Interspeech 2019}, pp.\
  146--150. {ISCA}, 2019.

\bibitem[Conneau et~al.(2020)Conneau, Baevski, Collobert, Mohamed, and
  Auli]{DBLP:journals/corr/abs-2006-13979}
Conneau, A., Baevski, A., Collobert, R., Mohamed, A., and Auli, M.
\newblock Unsupervised cross-lingual representation learning for speech
  recognition.
\newblock \emph{CoRR}, abs/2006.13979, 2020.

\bibitem[Devlin et~al.(2019)Devlin, Chang, Lee, and
  Toutanova]{DBLP:conf/naacl/DevlinCLT19}
Devlin, J., Chang, M., Lee, K., and Toutanova, K.
\newblock {BERT:} pre-training of deep bidirectional transformers for language
  understanding.
\newblock In Burstein, J., Doran, C., and Solorio, T. (eds.), \emph{{NAACL-HLT}
  2019}, pp.\  4171--4186. Association for Computational Linguistics, 2019.

\bibitem[Ghoshal et~al.(2013)Ghoshal, Swietojanski, and
  Renals]{ghoshal2013multilingual}
Ghoshal, A., Swietojanski, P., and Renals, S.
\newblock Multilingual training of deep neural networks.
\newblock In \emph{2013 IEEE International Conference on Acoustics, Speech and
  Signal Processing}, pp.\  7319--7323. IEEE, 2013.

\bibitem[Graves(2012)]{Graves-RNNSeqTransduction}
Graves, A.
\newblock Sequence transduction with recurrent neural networks.
\newblock \emph{CoRR}, abs/1211.3711, 2012.

\bibitem[Graves et~al.(2006)Graves, Fern{\'a}ndez, Gomez, and
  Schmidhuber]{graves2006connectionist}
Graves, A., Fern{\'a}ndez, S., Gomez, F., and Schmidhuber, J.
\newblock Connectionist temporal classification: labelling unsegmented sequence
  data with recurrent neural networks.
\newblock In \emph{ICML}, pp.\  369--376, 2006.

\bibitem[Hernandez et~al.(2018)Hernandez, Nguyen, Ghannay, Tomashenko, and
  Est{\`e}ve]{hernandez2018ted}
Hernandez, F., Nguyen, V., Ghannay, S., Tomashenko, N., and Est{\`e}ve, Y.
\newblock Ted-lium 3: twice as much data and corpus repartition for experiments
  on speaker adaptation.
\newblock In \emph{International Conference on Speech and Computer}, pp.\
  198--208. Springer, 2018.

\bibitem[Huang et~al.(2013)Huang, Li, Yu, Deng, and Gong]{huang2013cross}
Huang, J.-T., Li, J., Yu, D., Deng, L., and Gong, Y.
\newblock Cross-language knowledge transfer using multilingual deep neural
  network with shared hidden layers.
\newblock In \emph{2013 IEEE International Conference on Acoustics, Speech and
  Signal Processing}, pp.\  7304--7308. IEEE, 2013.

\bibitem[Huang et~al.(2014)Huang, Yu, Liu, and
  Gong]{DBLP:conf/interspeech/HuangYLG14a}
Huang, Y., Yu, D., Liu, C., and Gong, Y.
\newblock Multi-accent deep neural network acoustic model with accent-specific
  top layer using the kld-regularized model adaptation.
\newblock In Li, H., Meng, H.~M., Ma, B., Chng, E., and Xie, L. (eds.),
  \emph{{INTERSPEECH} 2014}, pp.\  2977--2981. {ISCA}, 2014.

\bibitem[Jang et~al.(2016)Jang, Gu, and Poole]{jang2016categorical}
Jang, E., Gu, S., and Poole, B.
\newblock Categorical reparameterization with gumbel-softmax.
\newblock \emph{arXiv preprint arXiv:1611.01144}, 2016.

\bibitem[Joshi et~al.(2020)Joshi, Zhao, Mehta, Kumar, and
  Li]{DBLP:conf/interspeech/JoshiZMKL20}
Joshi, V., Zhao, R., Mehta, R.~R., Kumar, K., and Li, J.
\newblock Transfer learning approaches for streaming end-to-end speech
  recognition system.
\newblock In Meng, H., Xu, B., and Zheng, T.~F. (eds.), \emph{Interspeech
  2020}, pp.\  2152--2156. {ISCA}, 2020.

\bibitem[Katzner \& Miller(2002)Katzner and Miller]{katzner2002languages}
Katzner, K. and Miller, K.
\newblock \emph{The languages of the world}.
\newblock Routledge, 2002.

\bibitem[Kawakami et~al.(2020)Kawakami, Wang, Dyer, Blunsom, and van~den
  Oord]{DBLP:conf/emnlp/KawakamiWDBO20}
Kawakami, K., Wang, L., Dyer, C., Blunsom, P., and van~den Oord, A.
\newblock Learning robust and multilingual speech representations.
\newblock In Cohn, T., He, Y., and Liu, Y. (eds.), \emph{{EMNLP} 2020}, pp.\
  1182--1192. Association for Computational Linguistics, 2020.

\bibitem[Knill et~al.(2014)Knill, Gales, Ragni, and
  Rath]{DBLP:conf/interspeech/KnillGRR14}
Knill, K., Gales, M. J.~F., Ragni, A., and Rath, S.~P.
\newblock Language independent and unsupervised acoustic models for speech
  recognition and keyword spotting.
\newblock In Li, H., Meng, H.~M., Ma, B., Chng, E., and Xie, L. (eds.),
  \emph{{INTERSPEECH} 2014}, pp.\  16--20. {ISCA}, 2014.

\bibitem[Kunze et~al.(2017)Kunze, Kirsch, Kurenkov, Krug, Johannsmeier, and
  Stober]{DBLP:conf/rep4nlp/KunzeKKKJS17}
Kunze, J., Kirsch, L., Kurenkov, I., Krug, A., Johannsmeier, J., and Stober, S.
\newblock Transfer learning for speech recognition on a budget.
\newblock In Blunsom, P., Bordes, A., Cho, K., Cohen, S.~B., Dyer, C.,
  Grefenstette, E., Hermann, K.~M., Rimell, L., Weston, J., and Yih, S. (eds.),
  \emph{ACL 2017 workshop}, pp.\  168--177. Association for Computational
  Linguistics, 2017.
\newblock \doi{10.18653/v1/w17-2620}.

\bibitem[Li et~al.(2014)Li, Deng, Gong, and Haeb-Umbach]{li2014overview}
Li, J., Deng, L., Gong, Y., and Haeb-Umbach, R.
\newblock An overview of noise-robust automatic speech recognition.
\newblock \emph{IEEE/ACM Transactions on Audio, Speech, and Language
  Processing}, 22\penalty0 (4):\penalty0 745--777, 2014.

\bibitem[Li et~al.(2020)Li, Wu, Gaur, Wang, Zhao, and Liu]{li2020comparison}
Li, J., Wu, Y., Gaur, Y., Wang, C., Zhao, R., and Liu, S.
\newblock On the comparison of popular end-to-end models for large scale speech
  recognition.
\newblock \emph{arXiv preprint arXiv:2005.14327}, 2020.

\bibitem[Ling et~al.(2020{\natexlab{a}})Ling, Liu, Salazar, and
  Kirchhoff]{DBLP:conf/icassp/LingLSK20}
Ling, S., Liu, Y., Salazar, J., and Kirchhoff, K.
\newblock Deep contextualized acoustic representations for semi-supervised
  speech recognition.
\newblock In \emph{{ICASSP} 2020}, pp.\  6429--6433. {IEEE},
  2020{\natexlab{a}}.

\bibitem[Ling et~al.(2020{\natexlab{b}})Ling, Salazar, Liu, Kirchhoff, and
  Amazon]{ling2020bertphone}
Ling, S., Salazar, J., Liu, Y., Kirchhoff, K., and Amazon, A.
\newblock Bertphone: Phonetically-aware encoder representations for
  utterance-level speaker and language recognition.
\newblock In \emph{Proc. Odyssey 2020 The Speaker and Language Recognition
  Workshop}, pp.\  9--16, 2020{\natexlab{b}}.

\bibitem[Liu et~al.(2020)Liu, Yang, Chi, Hsu, and
  Lee]{DBLP:conf/icassp/LiuYCHL20}
Liu, A.~T., Yang, S., Chi, P., Hsu, P., and Lee, H.
\newblock Mockingjay: Unsupervised speech representation learning with deep
  bidirectional transformer encoders.
\newblock In \emph{{ICASSP} 2020}, pp.\  6419--6423. {IEEE}, 2020.

\bibitem[Oord et~al.(2018)Oord, Li, and Vinyals]{oord2018representation}
Oord, A. v.~d., Li, Y., and Vinyals, O.
\newblock Representation learning with contrastive predictive coding.
\newblock \emph{arXiv preprint arXiv:1807.03748}, 2018.

\bibitem[Ott et~al.(2019)Ott, Edunov, Baevski, Fan, Gross, Ng, Grangier, and
  Auli]{ott2019fairseq}
Ott, M., Edunov, S., Baevski, A., Fan, A., Gross, S., Ng, N., Grangier, D., and
  Auli, M.
\newblock fairseq: A fast, extensible toolkit for sequence modeling.
\newblock \emph{arXiv preprint arXiv:1904.01038}, 2019.

\bibitem[Panayotov et~al.(2015)Panayotov, Chen, Povey, and
  Khudanpur]{panayotov2015librispeech}
Panayotov, V., Chen, G., Povey, D., and Khudanpur, S.
\newblock Librispeech: an asr corpus based on public domain audio books.
\newblock In \emph{ICASSP}, pp.\  5206--5210. IEEE, 2015.

\bibitem[Pascual et~al.(2019)Pascual, Ravanelli, Serr{\`{a}}, Bonafonte, and
  Bengio]{DBLP:conf/interspeech/PascualRSBB19}
Pascual, S., Ravanelli, M., Serr{\`{a}}, J., Bonafonte, A., and Bengio, Y.
\newblock Learning problem-agnostic speech representations from multiple
  self-supervised tasks.
\newblock In Kubin, G. and Kacic, Z. (eds.), \emph{Interspeech 2019, 20th
  Annual Conference of the International Speech Communication Association,
  Graz, Austria, 15-19 September 2019}, pp.\  161--165. {ISCA}, 2019.
\newblock \doi{10.21437/Interspeech.2019-2605}.
\newblock URL \url{https://doi.org/10.21437/Interspeech.2019-2605}.

\bibitem[Ravanelli et~al.(2020)Ravanelli, Zhong, Pascual, Swietojanski,
  Monteiro, Trmal, and Bengio]{DBLP:conf/icassp/RavanelliZPSMTB20}
Ravanelli, M., Zhong, J., Pascual, S., Swietojanski, P., Monteiro, J., Trmal,
  J., and Bengio, Y.
\newblock Multi-task self-supervised learning for robust speech recognition.
\newblock In \emph{2020 {IEEE} International Conference on Acoustics, Speech
  and Signal Processing, {ICASSP} 2020, Barcelona, Spain, May 4-8, 2020}, pp.\
  6989--6993. {IEEE}, 2020.
\newblock \doi{10.1109/ICASSP40776.2020.9053569}.
\newblock URL \url{https://doi.org/10.1109/ICASSP40776.2020.9053569}.

\bibitem[Rivi{\`{e}}re et~al.(2020)Rivi{\`{e}}re, Joulin, Mazar{\'{e}}, and
  Dupoux]{DBLP:conf/icassp/RiviereJMD20}
Rivi{\`{e}}re, M., Joulin, A., Mazar{\'{e}}, P., and Dupoux, E.
\newblock Unsupervised pretraining transfers well across languages.
\newblock In \emph{{ICASSP} 2020}, pp.\  7414--7418. {IEEE}, 2020.

\bibitem[Sadhu et~al.(2021)Sadhu, He, Huang, Mallidi, Wu, Rastrow, Stolcke,
  Droppo, and Maas]{DBLP:journals/corr/abs-2103-08393}
Sadhu, S., He, D., Huang, C., Mallidi, S.~H., Wu, M., Rastrow, A., Stolcke, A.,
  Droppo, J., and Maas, R.
\newblock Wav2vec-c: {A} self-supervised model for speech representation
  learning.
\newblock \emph{CoRR}, abs/2103.08393, 2021.
\newblock URL \url{https://arxiv.org/abs/2103.08393}.

\bibitem[Schneider et~al.(2019)Schneider, Baevski, Collobert, and
  Auli]{DBLP:conf/interspeech/SchneiderBCA19}
Schneider, S., Baevski, A., Collobert, R., and Auli, M.
\newblock wav2vec: Unsupervised pre-training for speech recognition.
\newblock In Kubin, G. and Kacic, Z. (eds.), \emph{Interspeech 2019}, pp.\
  3465--3469. {ISCA}, 2019.

\bibitem[Talnikar et~al.(2020)Talnikar, Likhomanenko, Collobert, and
  Synnaeve]{DBLP:journals/corr/abs-2011-00093}
Talnikar, C., Likhomanenko, T., Collobert, R., and Synnaeve, G.
\newblock Joint masked {CPC} and {CTC} training for {ASR}.
\newblock \emph{CoRR}, abs/2011.00093, 2020.
\newblock URL \url{https://arxiv.org/abs/2011.00093}.

\bibitem[Vaswani et~al.(2017)Vaswani, Shazeer, Parmar, Uszkoreit, Jones, Gomez,
  Kaiser, and Polosukhin]{DBLP:conf/nips/VaswaniSPUJGKP17}
Vaswani, A., Shazeer, N., Parmar, N., Uszkoreit, J., Jones, L., Gomez, A.~N.,
  Kaiser, L., and Polosukhin, I.
\newblock Attention is all you need.
\newblock In Guyon, I., von Luxburg, U., Bengio, S., Wallach, H.~M., Fergus,
  R., Vishwanathan, S. V.~N., and Garnett, R. (eds.), \emph{Advances in Neural
  Information Processing Systems 30: Annual Conference on Neural Information
  Processing Systems 2017, December 4-9, 2017, Long Beach, CA, {USA}}, pp.\
  5998--6008, 2017.

\bibitem[Wang et~al.(2020)Wang, Shi, Zhang, Wu, Chan, Yeh, and
  Xiao]{wang2020transformer}
Wang, Y., Shi, Y., Zhang, F., Wu, C., Chan, J., Yeh, C.-F., and Xiao, A.
\newblock Transformer in action: a comparative study of transformer-based
  acoustic models for large scale speech recognition applications.
\newblock \emph{arXiv preprint arXiv:2010.14665}, 2020.

\bibitem[Watanabe et~al.(2018)Watanabe, Hori, Karita, Hayashi, Nishitoba, Unno,
  Soplin, Heymann, Wiesner, Chen, et~al.]{watanabe2018espnet}
Watanabe, S., Hori, T., Karita, S., Hayashi, T., Nishitoba, J., Unno, Y.,
  Soplin, N. E.~Y., Heymann, J., Wiesner, M., Chen, N., et~al.
\newblock Espnet: End-to-end speech processing toolkit.
\newblock \emph{arXiv preprint arXiv:1804.00015}, 2018.

\bibitem[Xiong et~al.(2016)Xiong, Droppo, Huang, Seide, Seltzer, Stolcke, Yu,
  and Zweig]{xiong2016achieving}
Xiong, W., Droppo, J., Huang, X., Seide, F., Seltzer, M., Stolcke, A., Yu, D.,
  and Zweig, G.
\newblock Achieving human parity in conversational speech recognition.
\newblock \emph{arXiv preprint arXiv:1610.05256}, 2016.

\end{thebibliography}
\bibliographystyle{icml2021}



\end{document}